\documentclass{article}

\usepackage{arxiv}

\usepackage[utf8]{inputenc}	
\usepackage[T1]{fontenc}		
\usepackage{hyperref}			
\usepackage{url} 					
\usepackage{booktabs}			
\usepackage{amsfonts} 			
\usepackage{nicefrac}       		
\usepackage{microtype}      	
\usepackage{lipsum}

\usepackage{graphicx}
\usepackage{lineno}
\usepackage{natbib}
\usepackage{hyperref}
\usepackage{epsfig}
\usepackage{subfigure}
\usepackage{calc}
\usepackage{amssymb}
\usepackage{amstext}
\usepackage{amsmath}
\usepackage{mathtools}
\usepackage{multicol}
\usepackage{etoolbox}
\usepackage{enumitem}
\usepackage{diagbox}
\usepackage{float}
\usepackage{array}
\usepackage{adjustbox}

\modulolinenumbers[5]
\graphicspath{{./Figures/}}

\newcommand{\fold}[1]{\mathcal{#1}}

\DeclarePairedDelimiter{\ceil}{\lceil}{\rceil}

\hypersetup{
  pdfauthor = {},
  pdftitle = {},
  pdfsubject = {},
  pdfkeywords = {},
  colorlinks=true,
  linkcolor= blue,
  citecolor= blue,
  pageanchor=true,
  urlcolor = blue,
  plainpages = false,
  linktocpage
}

\newcommand{\secref}[1]{\autoref{sec:#1}}
\newcommand{\figref}[1]{\autoref{fig:#1}}
\newcommand{\eqnref}[1]{\autoref{eqn:#1}}
\newcommand{\tabref}[1]{\autoref{tab:#1}}

\title{Supporting Optimal Phase Space Reconstructions Using Neural Network Architecture for Time Series Modeling}

\author{Lucas~de~C.~Pagliosa, Alexandru~C.~Telea and~Rodrigo~F.~Mello
\thanks{L. Pagliosa and A. Telea were with the Faculty of Science and Engineering, University of Groningen, Nijenborgh 4, 9747 AG, Groningen, The Netherlands e-mail: l.de.carvalho.pagliosa@rug.nl, a.c.telea@rug.nl.}
\thanks{R. Mello was with the Institute of Mathematics and Computer Science, University of S\~ao Paulo, 400, 13566-590, S\~ao Carlos, Brazil e-mail: mello@icmc.usp.br.}}

\begin{document}
\maketitle

\begin{abstract}
The reconstruction of phase spaces is an essential step to analyze time series according to Dynamical System concepts. A regression performed on such spaces unveils the relationships among system states from which we can derive their generating rules, that is, the most probable set of functions responsible for generating observations along time. In this sense, most approaches rely on Takens' embedding theorem to unfold the phase space, which requires the embedding dimension and the time delay. Moreover, although several methods have been proposed to empirically estimate those parameters, they still face limitations due to their lack of consistency and robustness, which has motivated this paper. As an alternative, we here propose an artificial neural network with a forgetting mechanism to implicitly learn the phase spaces properties, whatever they are. Such network trains on forecasting errors and, after converging, its architecture is used to estimate the embedding parameters. Experimental results confirm that our approach is either as competitive as or better than most state-of-the-art strategies while revealing the temporal relationship among time-series observations.
\end{abstract}



\section{Introduction}
\label{sec:introduction}

Time-series analyses has become a key instrument for the evaluation of continuously collected data in several domains such as Medicine, Physics and Statistics~\citep{firmino:2014, box:15:book}. Such analysis generally involves the creation of a model (a regression function or a classifier, for instance) that usually leads to inconsistent results when built over raw data, specially if it contains chaotic behavior~\citep{brock:book:92}. In order to reach more reliable results, an alternative is to study time-series trajectories in the phase space, as proposed by the area of Dynamical Systems~\citep{ott:book:02, alligood:1996}. Besides leading to more robust models, the phase space also allows the inference of other important measures, such as the correlation dimension~\citep{grassberger:physicaD:83,mandelbrot:1977, theiler:josa:90, clark:llj:90, ding:physicaD:93} and the Lyapunov exponent~\citep{sano:prl:85, kantz:book:04}, which support further analyses in modeling.

\begin{figure}[htb]
\centering
\includegraphics[width=.60\linewidth]{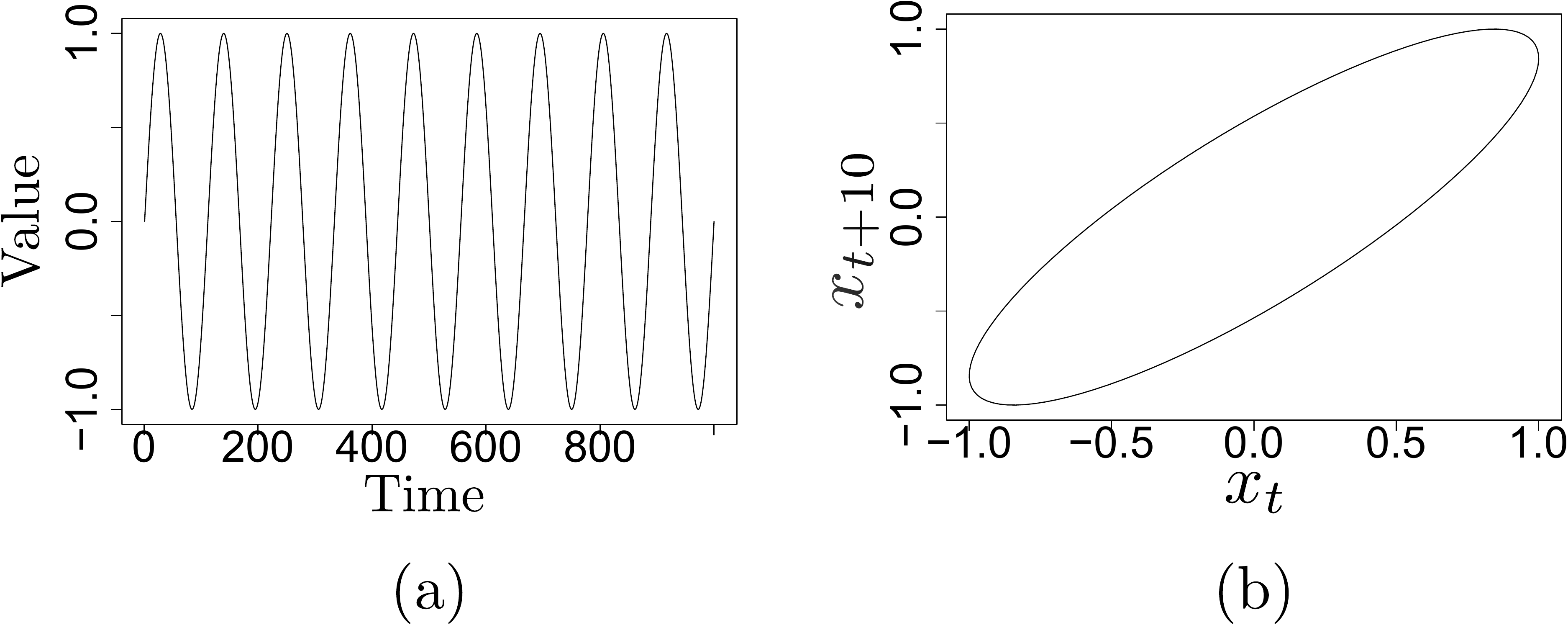}
\caption{The phase space of a sinusoidal function \textbf{(a)} is known to be sufficiently unfolded using the embedding dimension $m = 2$ and the time delay $\tau = 1$. However, for visualization purposes, we show such embedding with $\tau = 10$, as increasing the time delay increases the radius of the ellipse \textbf{(b)}.}
\label{fig:sine-example}
\end{figure}

In this context, Takens' embedding theorem~\citep{takens:1981} is one of the most used methods in the literature to reconstruct phase spaces from time series~\citep{ravindra:1998}. Such method relies on two parameters known as \emph{embedding dimension} $m$ and \emph{time delay} $\tau$ (see \figref{sine-example}) that, although Takens proved an arbitrary $\tau$ can be used given $m$ is sufficiently large, the minimum-but-sufficient (from now on denoted as \emph{optimal}) set of embedding parameters is desirable either to optimize phase-space computations as to better understand the analyzed phenomenon. In this context, several methods based on entropy~\citep{han:prl:2012}, fractal dimensions~\citep{theiler:josa:90} and/or nearest neighbors~\citep{kennel:1992} were proposed to guide the estimation of optimal embeddings. Despite contributions, those methods often present limitations such as lack of robustness against ground-truth dataset, sensitiveness to variations on the search space, and unclear heuristics involved in the selection of the model hyperparameters. Among other reasons, those problems derive from inconsistencies related to the curse of dimensionality~\citep{chen:book:09}, the number of employed nearest neighbors and the presence of noise. Thus, it is difficulty to define (and rely on) a set of specific properties to look at when searching for the optimal phase space, which is, as matter of fact, what most of state-of-the-art methods propose.

In attempt to improve those topics, we propose a multilayer feedforward neural network with backpropagation~\citep{delashmit:2005} to implicitly capture phase-space nonlinearities of any type. By performing learning with forgetting~\citep{ishikawa:1996}, we rely on a skeletal architecture (with many inhibited connection weights) to estimate the embedding parameters based on the converged network. In this context, we propose a simple but efficient technique for selecting the embedding parameters based on the convergence of network weights. In summary, our method differs from the literature in several aspects. When comparing to the traditional state-of-the-art~\citep{fraser:1986,kennel:1992,kember:physicsA:93,rosenstein:physicaD:94,rubinstein:book:07}, our estimation does not depend on the explicit definition of any phase-space features, which can drastically vary according to the dynamical system. On the other hand, when compared to similar neural-network approaches~\citep{karunasinghe:joh:06,manabe:neuro:07}, our method does not require expensive Monte Carlo simulations for parameter setup and it is much simpler and faster. Moreover, we improve the literature with the following contributions:

\begin{enumerate}[label=R\arabic*, leftmargin = 1.1cm]
    \item Few user-defined parameters and settings involved in the embedding estimation process;
    \item Low sensitivity and complexity in performing searches on the space of parameters, a.k.a. search space;
    \item Robust validation against ground-truth datasets.
\end{enumerate}

The remaining of this paper is structured as follows: \secref{background} introduces background information and notations; \secref{related-work} briefly discusses the related work; \secref{method} details our proposed method; \secref{experiments} presents experimental results on both synthetic and real-world datasets; \secref{conclusion} draws concluding remarks and future directions.

\section{Background}
\label{sec:background}

A dynamical system ${\fold{S}^d = \{p_0, \cdots, p_t\}}$ is composed of a set of $d$-dimensional states in the form ${p_k = (p_{k1}, p_{k2}, \cdots, p_{kd})}$ that, driven by a generating rule $R(\cdot)$, models the behavior of a certain phenomenon as a function of its states trajectories (evolution in time) such that:
\begin{equation}
R : \fold{S}^d \rightarrow \fold{S}^d,
\end{equation}
where $d$ is the system degree of freedom, \emph{i.e.}, the number of dimensions or variables describing $R(\cdot)$. Having enough states to simulate all possible dynamics of $R(\cdot)$, $\fold{S}^d$ is referred to as a \emph{phase space}, which is represented by a potentially lower-dimensional manifold~\citep{lee:2003, tu:book:10} as the trajectories converge to its \emph{attractor}. When a system is described by such a phase space, the variable indicating time is no longer explicitly required for modeling~\citep{pagliosa:eswa:17}.

Therefore, one key advantage of the phase space is that it can be used to analyze how a phenomenon evolves along \emph{any} given period of time, therefore more robustly supporting (i) the finding of patterns (\emph{e.g.} cycles and trends); (ii) the forecasting of observations; and (iii) the correlation with different systems.

Moreover, a univariate time series $T_i$ can be seen as a sequence of $n$ observations:
\begin{equation}
T_i = \{x_0, \cdots, x_{n - 1}\}, ~ x_k \subset \mathbb{R},
\label{eqn:time-series}
\end{equation}
that, in our context, models the evolution of a single dimension $i \in [1, d]$ of $\fold{S}^d$. Due to the recurrent behavior of real-world processes, it is assumed that observations from other dimensions not just repeat among themselves but directly or indirectly affect others along time, such that $\fold{S}^d$ could be discovered from $T_i$ if the series suffers enough influences from other variables. According to Takens' embedding theorem~\citep{takens:1981}, $T_i$ can be embedded into an $m$-dimensional space diffeomorphic to $\fold{S}^d$ if phase states are in the Standard Embedding Vector (SEV) form:
\begin{equation}
p_t = (x_t, x_{t + \tau}, \cdots x_{t + (m - 1)\tau}),
\label{eqn:sev}
\end{equation}
where $\tau$ (a.k.a. time delay) describes the recurrence time in $\fold{S}^d$. Moreover, despite such variable could assume different values along dimensions, it is commonly represented as a single value (the average or the higher time delay, for instance).

\begin{figure}[htb]
\centering
\includegraphics[width=.75\linewidth]{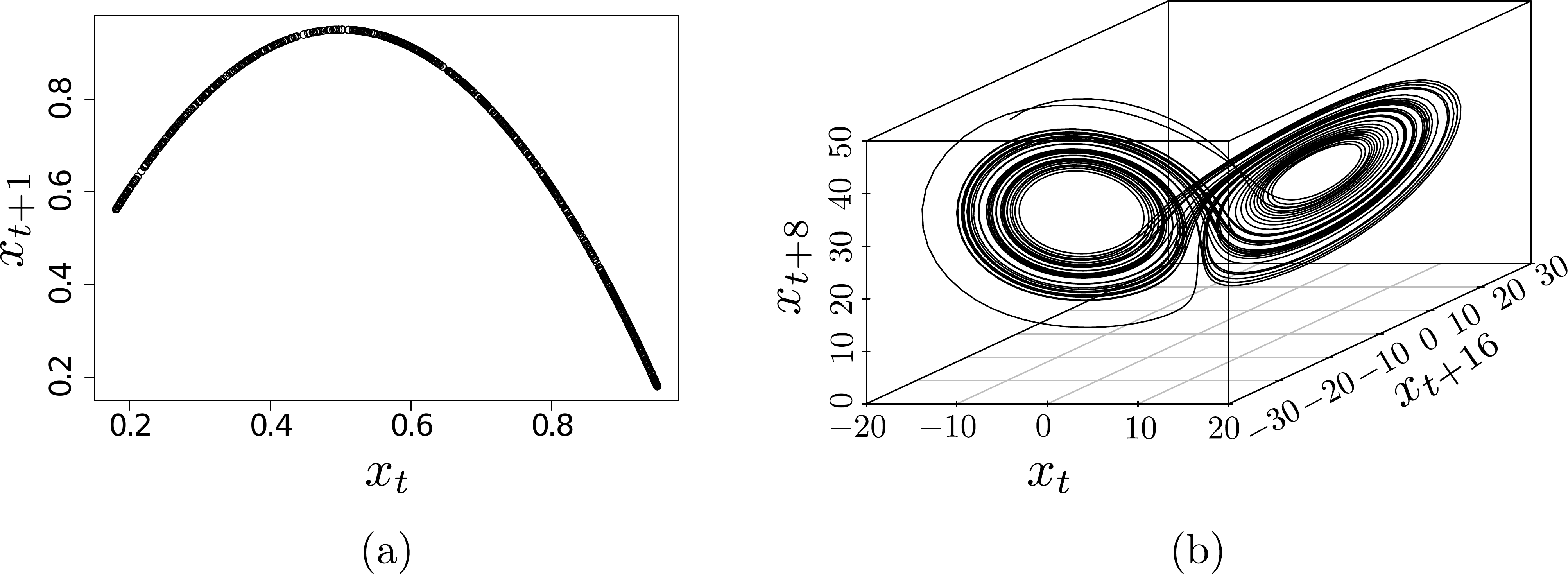}
\caption{Logistic map \textbf{(a)} and Lorenz system \textbf{(b)} phase spaces.}
\label{fig:embeddings-example}
\end{figure}

\figref{embeddings-example} shows the phase spaces for the well-known Logistic~\citep{may:nature:1976,robledo:2007} and Lorenz~\citep{tucker:1999} time series\footnote{Details about all the datasets used in this manuscript are found in~\citep{pagliosa:eswa:17}.}, each of those embedded using (${m = 2, \tau = 1}$) and (${m = 3, \tau = 8}$), respectively.
As it can be seen from such a figure, states in the phase space compose a well-formed structure as they are bounded by the attractor, what could suggest the usage of entropy~\citep{han:prl:2012}, fractal dimensions~\citep{theiler:josa:90} and/or nearest neighbors~\citep{kennel:1992} to guide the estimation of optimal embeddings. However, despite those measurements describe important aspects of phase spaces~\citep{alligood:1996}, they often lead to different results according to the methodology employed to compute them, which derives from inconsistencies related to the partition of the space into bins, the computation of probabilities, the curse of dimensionality~\citep{chen:book:09}, the number of nearest neighbors and the presence of noise. Moreover, some of those measurements are not suitable to distinguish phase spaces as they are not unique descriptors (different embeddings may have similar values of entropy/fractal dimension). All those issues contribute to the difficulty in define (and rely on) a set of specific properties to look at when searching for the optimal phase space, which is, as matter of fact, what most of state-of-the-art methods propose (see \tabref{comparison-results}).

In order to exemplify the benefits and problems of dealing with phase spaces, we go further into the generating rule of the Logistic map, defined as:
\begin{equation}
x_{t+1} = r x_t (1-x_t).
\label{eqn:logistic-map}
\end{equation}
Such map models the growing rate of populations like humans and bacteria, and despite its chaotic behavior for $r = 3.8$, a second-order polynomial regression applied over \figref{embeddings-example}(a) is enough to get an approximation of \eqnref{logistic-map}. The main advantage is that, once we have the generating rule, all analyses become much simpler and more reliable rather than performing any other assessment on top of the time series itself~\citep{mello:book:18}.

\begin{figure}[htb]
\centering
\includegraphics[width=.65\linewidth]{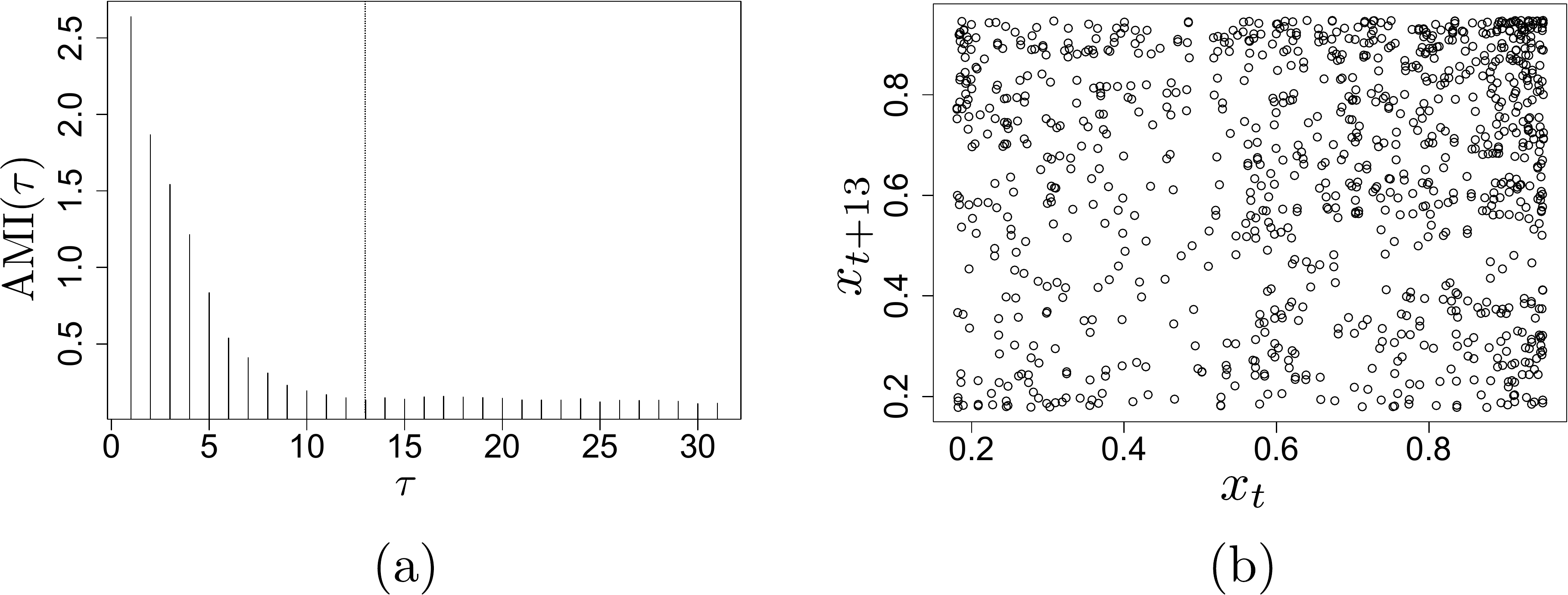}
\caption{\textbf{(a)} Estimation of the time delay for the Logistic map using the Auto-Mutual Information (AMI). As proposed in~\citep{fraser:1986}, $\tau$ is estimated based on the first local minimum reached after the initial lag equals to one. (\text{b}) Embedding of the Logistic map using $(m = 2, \tau = 13)$.}
\label{fig:ami-logistic}
\end{figure}

As a consequence, the quality of a phase space can only be measure (with $100$\% of certainty) if the generating rule is available for consultancy, and validations are usually driven on benchmark datasets. As an example, if one uses FNN and AMI to estimate $m$ and $\tau$ for the Logistic map, (those are among the most famous methods in the literature, see details in \secref{related-work}), the result is the pair (${m = 2, \tau = 13}$) -- see \figref{ami-logistic}(a)\footnote{For reproductive purpose: this figure was produced using the default options of the \texttt{mutual} function from the R package \texttt{tseriesChaos}~\citep{tsc:package:13}.}, what yields the phase space in \figref{ami-logistic}(b). As it can be noticed, the attractor structure is lost, resembling now a phase space from a random time series~\citep{alligood:1996}. No simple regression model would be capable of providing the generating rule of \eqnref{logistic-map} in such space.

When the generating rule is not known, however, a reasonable approach to validate the reconstructed phase space is to measure the quality of a regression model on this new found space, usually in terms of forecasting. It is worth to mention, however, that other researches might prefer to base their estimations of $m$ and $\tau$ on other measurements, such as the correlation dimension~\citep{grassberger:physicaD:83}. Further, despite we know there is no guarantee that the optimal embedding to forecast might not be the same to compute correlations, we have focused on the former due to practical reasons: forecasting lies among the main (if not the most) important goal while modeling time series.

\section{Related Work}
\label{sec:related-work}

Despite the importance of Takens' embedding theorem, that author has not provided any additional information on how to estimate the embedding parameters, only that a \emph{sufficient} dimension $m$ should be at least twice bigger than $d$ to properly unfold the phase space (although this is usually an overestimation). Moreover, several methods were proposed to estimate $m$ and $\tau$ under the assumption they are independent or bounded by the time-delay window ${t_w = (m - 1)\tau}$ \citep{albano:physicaD:91}. Next, we overview some key results in both of these topics, making the proper association with requirements R1--R3, introduced in \secref{introduction}. For a broader related work overview, refer to~\citep{otani:note:97}.

Early methods~\citep{albano:book:87, albano:physRevA:88, abarbanel:1993} used Autocorrelation Functions (ACFs) to estimate $\tau$, which have limited modeling abilities given only linear functions were employed. \cite{fraser:1986} tried to overcome those issues by using the first local minimum (from lag equals to $1$) of the nonlinear Auto-Mutual Information (AMI) function over different time delays (see \figref{ami-logistic}(a)). This simple approach respects R1 and R2 as it barely contains any parameters. Nonetheless,~\cite{martinerie:physreva:92} empirically observed that neither the ACF nor the AMI were consistent to estimate the time-delay window $t_w$ (and, as consequence, a bound for $\tau$), violating R3. 

\cite{kennel:1992} proposed the False Nearest Neighbors (FNN) method to estimate the optimal embedding dimension $m$. By using the time delay $\tau$ estimated using AMI, FNN reconstructs a time series using different dimensions while computing the index set of the $k$ nearest neighbors for each phase state. The best value for $m$ is defined as the one for which the fraction of nearest neighbors remains constant as the dimension increases. In spite of being simple and requiring an acceptable number of parameters (thus, satisfying R1), this method is very sensitive to the choice of $\tau$ and noise, counterposing R2 and R3.

\cite{rosenstein:physicaD:94} employed the Average Displacement (AD) measure to gauge the inverse relation between the redundancy error and the attractor expansion as a function of the time delay. They observed that AD increases until it reaches a plateau, indicating the attractor is sufficiently expanded. However, non-negligible errors are typically introduced while analyzing general systems~\citep{ma:feee:06}, what goes in disagreement with R3; similarly to FNN, this method involves Monte Carlo simulations~\citep{rubinstein:book:07} while scanning the space of parameters, failing R2.

The expansion of an attractor can also be described in terms of the spreading rate of its phase states, \emph{i.e.}, as function of its singular values. In this context, \cite{kember:physicsA:93} proposed the Singular Value Fraction (SVF) to estimate the time delay when the attractor is maximally spread out, which ideally should happen when all eigenvalues are equal. As this is unlikely to occur for real-world scenarios, the time delay $\tau$ yielding the minimum SVF was defined as the most adequate to represent the phase space. In summary, this method is simple and demands no parameters to compute, which satisfies R1 and R2. However, despite SVF shows consistent results for different dimensions, as recently reinforced by a modified version~\citep{chen:matec:16}, it may not properly work for attractors whose genii (number of voids in the manifold) is greater than $1$, thus it does not fully meet R3.

\cite{temujin:icassp:03} realized that a deterministic attractor should have a well-formed structure and, therefore, low entropy. Thus, they proposed the Entropy Ratio (ER), a method based on minimizing the ratio between the entropy from the phase spaces of the original series and a set of surrogates, providing a function similar to the Minimum Description Length~\citep{rissanen:1978}. In this scenario, R2 is not held as the method needs to reconstruct the phase space for all parameter combinations in order to assess the minimum ER. In addition, no consistency was achieved for such approach either (failing R3). As said before, the entropy is not a unique descriptor and might not be the best feature to characterize phase spaces. 

In spite of several studies involving the prediction of time series through the usage of neural-network models~\citep[etc.]{chakraborty:92, karunasinghe:joh:06, bhardwaj:book:10, han:ijcnn:13}, to the extent of our knowledge, only two of those approaches attempted to estimate $m$ and $\tau$. The first by~\cite{karunasinghe:joh:06} selected the set of embedding parameters over a densely-sampled range of values based on forecasting accuracies, which violates R2. In addition, their results were overestimated for ground-truth datasets, failing R3. 

The second approach, by~\cite{manabe:neuro:07}, proposed a more consistent strategy for estimating $m$ and $\tau$ without the need of exhaustive comparisons. They start using the FNN and AMI methods to define the maximum embedding bounds (MEB), \emph{i.e.}, the upper bound for $m$ and $\tau$ respectively, referred from now on as $m_{\max}$ and $\tau_{\max}$. The phase states were then constructed in the Provisional Embedding Vector (PEV) form, as follows:
\begin{equation}
p_t = (x_t, x_{t + 1}, \cdots x_{t + (m - 1)\tau}),
\label{eqn:pev}
\end{equation}
such that $|\text{PEV}| = {(m - 1)\tau + 1}$ observations are taken into account, differently from the $m$ observations in the SEV format (\eqnref{sev}). The vector in \eqnref{pev} is then used as input layer in a Multilayer Perceptron~\citep{delashmit:2005}, to later be propagated to the hidden layer (the number of neurons in this layer was not detailed by the authors), and next to a single output neuron to forecast $\rho$ steps ahead, in form:
\begin{equation}
f_\text{NN}(p_t) = x_{t + (m-1)\tau + \rho},
\label{eqn:manane-mapping}
\end{equation}
where ${f_\text{NN}: R^{PEV} \rightarrow R^1}$ is the neural-network predictive mapping, whose production involved learning with forgetting, hidden unit clarification, selective learning, and pruning heuristics to produces a final skeletal network. Thus, this approach required various thresholds and parameter settings that increased the modeling complexity and chances of overfitting (violating R1 and R2). After the network converges, embedding parameters were estimated directly from its architecture. As proposed, $m$ and $\tau$ were driven based on the most relevant (greater absolute magnitudes) weights connecting input-to-hidden layers. Finally, no test was performed on more complex datasets such as the Lorenz and R\"ossler systems, so that R3 was not fully covered.

In our point of view, the main contribution by~\cite{manabe:neuro:07} was to infer $m$ and $\tau$ without explicitly defining any phase-space features. Hence, phase-space inconsistencies such as expansion rate, noise, number of genus and redundancy are disregarded in their approach. On the other hand, the method is complex, it requires expensive Monte Carlo simulations for determining parameter values and, finally, it yields to different results due to the random weights initialization.

\section{The Proposed Method}
\label{sec:method}

In this section, we introduce our estimation method and compare it against~\cite{manabe:neuro:07} (referred to as MC), which is the most similar study found in the literature. Along the manuscript, we detail our approach differences and improvements in comparison to MC. Firstly, we describe our network architecture and its settings (\secref{neural-network}) to next discuss how $m$ and $\tau$ can be confidently inferred from this proposal (\secref{visual-inspection-of-ps-parameters}).

\subsection{Network Architecture and Settings}
\label{sec:neural-network}

\begin{figure}[htb]
\centering
\includegraphics[width=.50\linewidth]{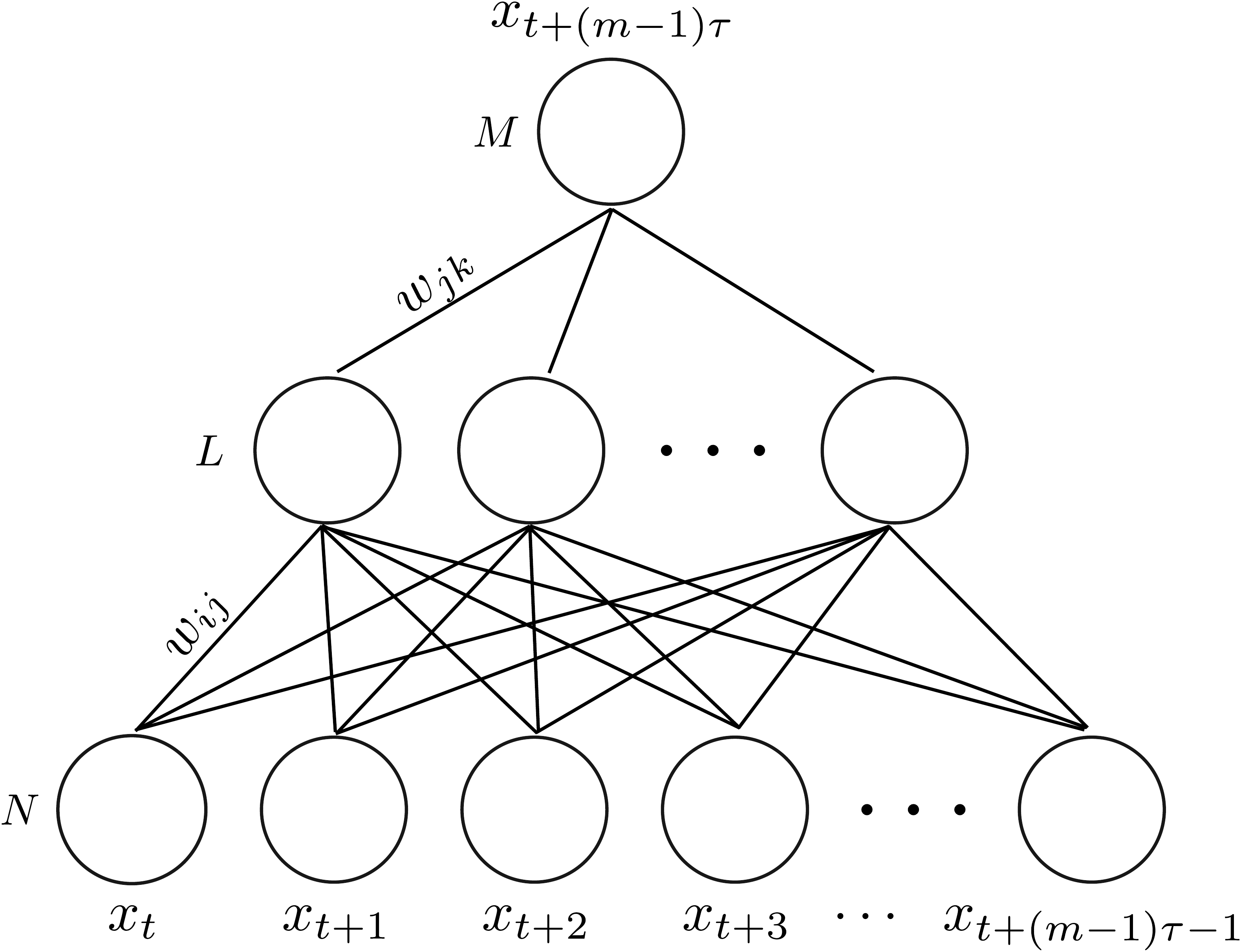}
\caption{Architecture of our three-layer neural network. Terms $w_{ij}$ and $w_{jk}$ represent input-to-hidden and hidden-to-output weights, respectively.}
\label{fig:neural-network-architecture}
\end{figure}

Our model is based on a fully-connected three-layer neural network trained using the backpropagation algorithm (\figref{neural-network-architecture}), implemented\footnote{The source code is available at~\url{https://github.com/pagliosa/neural-network}.} using the R statistical software~\citep{R:lang:08}. The triple $(N, L, M)$ represents the number of input, hidden, and output neurons, respectively. Similarly to MC, our architecture is based on PEV to forecast a single observation so that $M$ is always set equals to one. In contrast to MC (see \secref{related-work}), we restrict our input layer to $N = |\text{PEV}| - 1$ neurons and force the last PEV observation to define the class label to be used by the output layer, such that always $\rho = 1$ (\eqnref{manane-mapping}). Despite a small detail, such restriction is important to avoid overlearning without respecting the butterfly effect~\citep{brock:book:92}, \emph{i.e.}, when a predicted observation is fed in a recurrent fashion to the dataset to be used as a new query. In those cases, a recursive forecasting should be used as discussed in~\citep{pagliosa:eswa:17}.

In addition, we explicitly set $L = \ceil{\log(N)} + 1$ to probabilistically ensure the algorithm search space (a.k.a. bias) is in parsimony with the Bias-Variance Dilemma~\citep{geman:92:nc,vapnik:98:book, luxburg:11:book}. In other words, by logarithmically increasing $L$, we simultaneously avoid underfitting (the search space gets bigger and more functions can be used to fit data) and overfitting (it grows in a moderate pace based on the number of input neurons, which holds the model complexity).

Moreover, our architecture includes learning with forgetting by using the following cost function:
\begin{equation}
 C = \min_{p_t \in \fold{S}^d} \left( \sum_t E(p_t) + \lambda \sum_{e_{ij} \in K} |w_{ij}| \right),
\label{eqn:lwf}
\end{equation}
in which $E(p_t)$ is the error function given an input $p_t$, $w_{ij}$ is the weight for edge $e_{ij}$, and $K$ is the set of all $(N \times L)$ input-to-hidden network edges. Parameter $\lambda$ adjusts the trade off between weight minimization and accuracy performance. However, setting an ``adequate'' variable for $\lambda$, generically known as regularization term, is usually a problem by itself that requires either trial-and-error procedures or knowledge from a specialist. In matter of fact, the MC method adopted the first approach once it sets $\lambda$ based on the relative normalized score (RNS) and Monte Carlo simulations~\citep{landau:05:book,rubinstein:book:07}. Thus, after testing several values for $\lambda$, MC chose the value that led to the smallest prediction error. However, our experiments suggest that this step can be heuristically improved: despite learning is \emph{always} important to any Machine Learning approach, in our case we are strictly dependent on the final network weights distribution (see \secref{visual-inspection-of-ps-parameters} for details). Thus, by setting a``strong'' forgetting threshold ${\lambda = 10^{-3}}$, we exchange a bit of forecasting accuracy to force the neural network to converge to a skeletal architecture, such that $m$ and $\tau$ can be systematically derived. As we later show in \secref{experiments}, this has led to consistent and relevant results according to our experiments.

Moreover, our model simplifies MC as it does not depend on hidden unit clarification, selective forgetting or pruning heuristics. By removing such elements, the training stage becomes faster and more robust once it required smaller search spaces while being less prone to overfitting. Training was performed until cost $C$ (\eqnref{lwf}) reached a predefined threshold $C_{\max}$ or a maximum number of epochs $g$ (in our experiments, those parameters were set as $0.001$ and $500$, respectively). \tabref{network-settings} lists the settings and compare them against MC, including the momentum rate $\alpha$ and the step size $\eta$, both employed by the gradient descent method.
\begin{table}[htb]
\renewcommand\arraystretch{1.2}
\setlength{\tabcolsep}{3pt}
\newcolumntype{L}{>{\centering\arraybackslash}m{.5cm}}
\centering
\caption{Network settings of our model and comparison with MC ($n/a$ refers to missing information).}
\label{tab:network-settings}
 \begin{tabular}{|c|c|c|}
    \hline
    \diagbox{Parameter}{Method} & MC & Ours \\ \hline
    Number of input neurons $N$ & $|\text{PEV}|$ & $|\text{PEV}| - 1$ \\ \hline
    Number of hidden neurons $L$ & $n/a$ & $\log N + 1$ \\ \hline
    Number of output neurons $M$ & $1$ & $1$ \\ \hline
    Step size $\eta$ & $0.1$ & $0.1$ \\ \hline
    Momentum rate $\alpha$ & $0.2$ & $0.2$ \\ \hline
    Forgetting parameter $\lambda$ & set by RNS & $0.001$ \\ \hline
    Number of epochs $g$ & $50000$ & $500$ \\ \hline
    Maximal error tolerance $C_{\max}$ & $n/a$ & 0.001 \\ \hline
    Interval of random weights & $n/a$ & [-0.1, 0.1] \\ \hline
\end{tabular}
\end{table}

Our only free parameter is the size of the input layer $N$, defined as the length of the PEV. In contrast, MC defines it based on the estimations given by the FNN and AMI methods. As we understand, this approach tends to work well only when FNN and AMI do not overestimate embedding parameters, otherwise \eqnref{pev} would give a too long vector, increasing the complexity of the model. Conversely, in the case those methods underestimate $m$ and $\tau$, the Provisional Embedding Vector (PEV) may be too short, resulting in a poor architecture and in not enough information to learn about the underlying phenomenon. Therefore, given the inconsistencies of FNN and AMI, we understand these methods can be used just to guide the user to find smaller-to-medium values for $N$ rather than to define it. In that sense, we analyze the impact of different search spaces (values for MEB) in \secref{lorenz}.

\subsection{Visual Inspection of Embedding Parameters}
\label{sec:visual-inspection-of-ps-parameters}

\begin{figure}[htb]
\centering
\includegraphics[width=.45\linewidth]{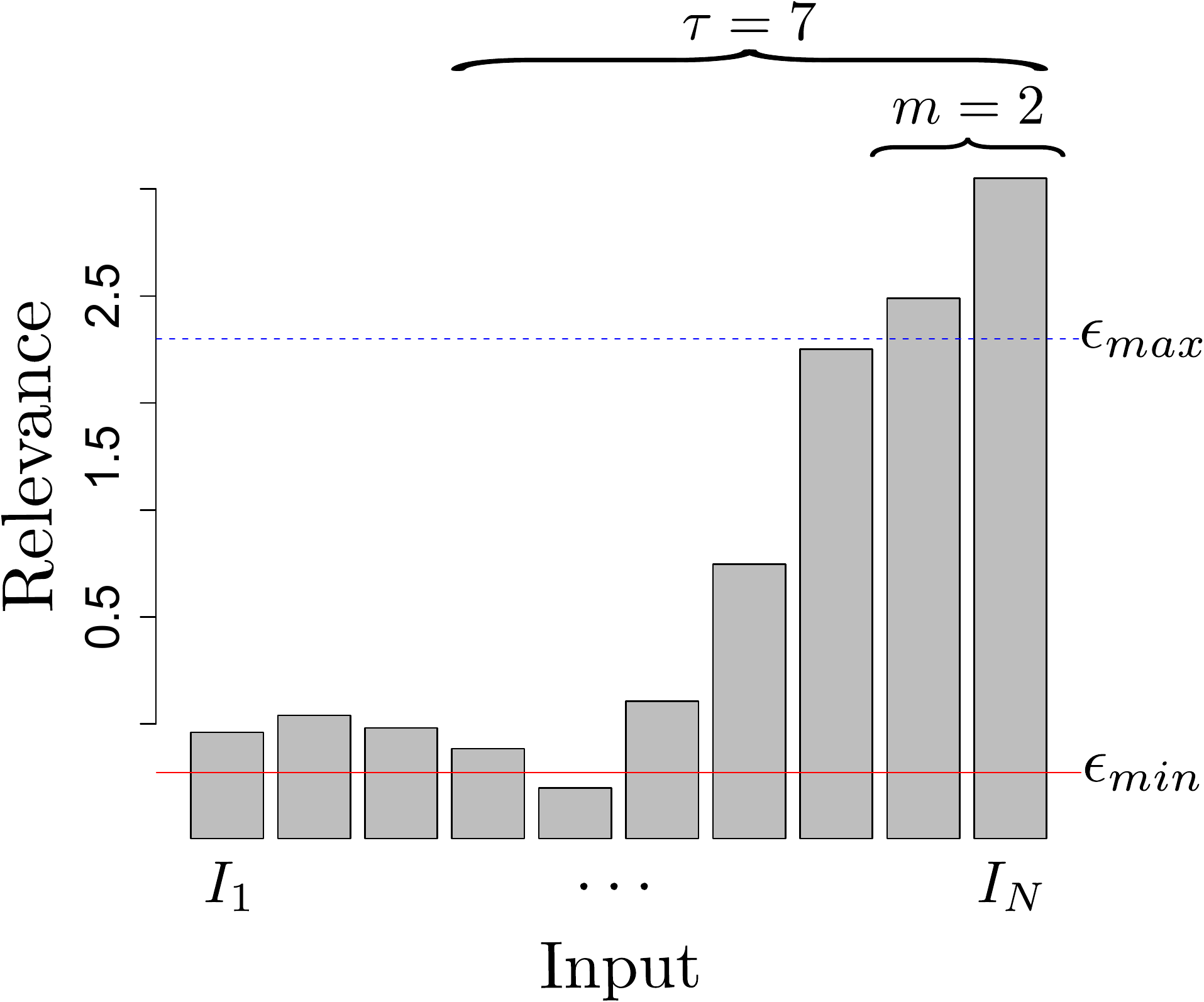}
\caption{Bar chart representing the relevance of input dimensions. Each bar corresponds to the sum $I_i$ of connection weights $w_{ij}$, given the input neuron $i$, to every hidden neuron identified with $j$. The dashed blue and solid red lines illustrate the thresholds $\epsilon_{\max}$ and $\epsilon_{\min}$ respectively, both used to determine the embedding parameters, which were set as ${(m,\tau)=(2,7)}$ in this example. For simplicity, indexes $I_1, \cdots, I_N$ are not shown in future plots.}
\label{fig:dimension-relevance}
\end{figure}

From a local perspective, each of the $N$ input neurons of our neural-network architecture corresponds to an observation of the PEV. From a global point of view, however, each input neuron can be seen as a \emph{dimension} of a representative basis. As our neural network is fully connected, we measured the relevance of each dimension $i \in [0, N]$ in terms of the sum $I_i = \sum^L_{j = 1} w_{ij}$, in which $w_{ij}$ is the weight associated with the connections between the $i^{th}$ input neuron to the $j^{th}$ hidden neuron. Such relevance can be depicted by a bar chart, in which the length of the $i^{th}$ bar maps the magnitude $I_i$ of a given input neuron $i$ (\figref{dimension-relevance}). 

We next use this bar chart to select the embedding parameters $m$ and $\tau$ inferred from our network model, as follows. Firstly, we consider all dimensions (at least two) whose relevance exceeds a quantile measure of $\epsilon_{\max} = 80\%$ over all $I_{1 \leq i \leq N}$, as relevant enough to represent parameter $m$. Secondly, the distance $|j - k| + 1$ results in the time delay $\tau$ which corresponds to the lag between the most and the least relevant terms $I_{j}$ and $I_{k}$, respectively. Notice $I_{k}$ is not simply associated with the smallest value, but with the least relevant dimension that lies above a minimum threshold of $\epsilon_{\min} = 10\%$ of $I_{j}$. In another words, the time delay is the first local minimum (from right to left) above $\epsilon_{\min}$. If no such delay is found, we set $\tau = 1$ as, in practice, it is the smallest possible value for the time delay. It is worth to mention that, despite $\epsilon_{\max}$ and $\epsilon_{\min}$ are free parameters that should be modified based on the search space (see \secref{lorenz} for details), we suggest to set them as $80\%$ and $10\%$, respectively, as those values usually represent a fair quantile/minimum value to describe general distributions. Indeed, they were enough to describe most of the benchmark datasets under our experimental analysis (\secref{experiments}). Moreover, the extreme cases for those parameters are such that $\epsilon_{\max}$ is greater enough to describe at least two relevant dimensions (as the minimum dimension a phase space can be embedded into is $m = 2$); and $\epsilon_{\min}$ is greater enough to describe at least one local minimum (assuming data is deterministic -- see \secref{sunspot} for discussion on stochastic data).

In summary, our method contributes to the related work in the sense that it does not require any specific definition on phase-space features (such as rate of expansion, number of false neighbors, entropy, etc.) to estimate the optimal $m$ and $\tau$. As we propose, the neural network learns those properties while training on forecasting errors, and we select the embedding parameters based on the final architecture. On the other hand, despite our article shares similarities with MC, we improve/differ from this method in the following topics: simplified training process; improved network architecture and settings; different approach to estimate $m$ and $\tau$ from the architecture; and experiments/discussion regarding variations on the search space and random initialization of neuron weights. 

\section{Experiments}
\label{sec:experiments}

We performed experiments to assess our method in light of the requirements R1--R3 (for more details, see \secref{introduction}). Next, we introduce the datasets used in the evaluation process (\secref{datasets}), while we discuss the obtained results and aspects of our proposal from \secref{logistic} to \secref{sunspot}. It is worth to mention that the ground-truth values for the given datasets were estimated according to one of the following cases:
\begin{itemize}
\item the generating rule $R(\cdot)$ of the dataset is known, hence embedding parameters can be obtained by trial-and-error after comparing the original and reconstructed phase spaces;
\item a specialist in the area of the respective dynamical system defined the phase space that best represents the dynamics of the studied phenomenon; or
\item the embedding dimensions and corresponding embedding parameters $(m, \tau)$ were estimated according to state-of-the-art methods which are well documented and accepted in the related literature.
\end{itemize}
Additionally, systems based on partial equations were solved using the sampling time $t_s = 0.01$ and $n = 1000$ observations, which have shown to be sufficient to preserve the dynamics of the unfolded phase space.

\subsection{Datasets and Methodology}
\label{sec:datasets}
In attempt to validate our method, we considered four benchmark datasets, namely Logistic map~\citep{robledo:2007}, H\'enon map~\citep{robledo:2007}, Lorenz system~\citep{tucker:1999}, and R\"ossler system~\citep{rossler:76:physics}. Those datasets were chosen because their respective generating rules $R(\cdot)$ are known, and their expected attractors can be fairly compared from the perspective of our approach. Additionally, we consider the Sunspot dataset~\citep{andrews:85:book} to support an empirical analysis based on real-world data. Finally, a discussion about how our method behaves while analyzing stochastic data, following a Gaussian distribution, is also performed.

The description and other details on those systems can be found in~\cite{pagliosa:eswa:17}. Thus, we do not detail those datasets but simply show in \tabref{comparison-results} their expected embedding parameters, as well as their respective predicted values by existing methods, whenever available. The embedding dimensions and time delays were defined as single or multiple possible values according to the extensive analysis provided in the related work~\citep[etc.]{rossler:76:physics, tucker:1999, robledo:2007}. It is also important to mention that the results obtained with FNN were only properly estimated after using the ground-truth time delay values, depicting a clear limitation. Whenever $\tau$ was computed using AMI, as usually performed in conjunction with FNN, the expected embedding dimension $m$ was hardly ever found.

All time series were normalized in range $[0, 1]$, such that our network weights were randomly initialized using $10\%$ of this range. In other words, rather than taking the typical weight range $[-1,1]$, as we suppose MC did once there is no additional information, we considered just the interval $[-0.1, 0.1]$ to bring solutions closer to the quasi-convex region of the squared-error surface as analyzed in~\citep{mello:book:18}. In practice, this initialization strategy was confirmed to provide better accuracy results than the most typical range of $[-1,1]$. Finally, we used a $5$-resampling validation criterion in all experiments, always taking $75\%$ of data for training and the remaining $25\%$ for testing.
\begin{table*}[ht]
\renewcommand\arraystretch{1.5}
\small
\centering
\caption{Comparison of embedding parameters ($m,\tau$). From left to right: datasets tested, ground truth (expected values according to the generating rule), results given by existing methods and ours. As one may notice, some methods either only estimate $m$ or $\tau$, whereas ER, MC, and ours estimate both. Terminology $n/a$ denotes datasets without a known ground truth (column 2) or which were not handled by MC (column 8).}
\label{tab:comparison-results}
\begin{adjustbox}{max width=\linewidth}
\begin{tabular}{|c|c|c|c|c|c|c|c|c|}
\hline
Series & Expected & AMI & FNN  & AD  & SVF  & ER & MC & Our method \\
 &  $(m,\tau)$ &  ($\tau$) &  ($m$) &  ($\tau$) &  ($\tau)$ &  ($m,\tau)$ & ($m,\tau)$ &  ($m,\tau)$\\
\hline
Logistic & (2--3, 1) & $13$ & $2$ & $3$ & $1$ & (2, 1) & $n/a$ & (2, 1)\\ \hline
H\'enon & (2--4, 1) & $12$ & $3$ & $3$ & $1$ & (3, 1) & (2--3, 1--5) & (2,1) \\ \hline
Lorenz & (2--3, 5--12) & $17$ & $2$ & $14$ & $55$ & (5, 1) & $n/a$ & (3, 8--12) \\ \hline
R\"ossler & (3, 5--12) & $13$ & $2$ & $11$ & $10$ & (5, 1) & $n/a$ & $(3,5)$\\ \hline
Sunspot & $n/a$ & $6$ & $3$ & $10$ & $59$ & (2, 1) & (2--4, 1--7) & (2, 1) \\ \hline
\end{tabular}
\end{adjustbox}
\end{table*}

\subsection{Logistic and H\'enon: Consistency along the five resamplings}
\label{sec:logistic}

One of the drawbacks of neural networks is the output of different results owing to the random weight initialization. While this aspect is less important for pure classification or regression tasks, it becomes crucial when information is extracted from the network architecture, as in our case (\secref{visual-inspection-of-ps-parameters}).

\begin{figure}[ht]
\centering
\includegraphics[width=.85\linewidth]{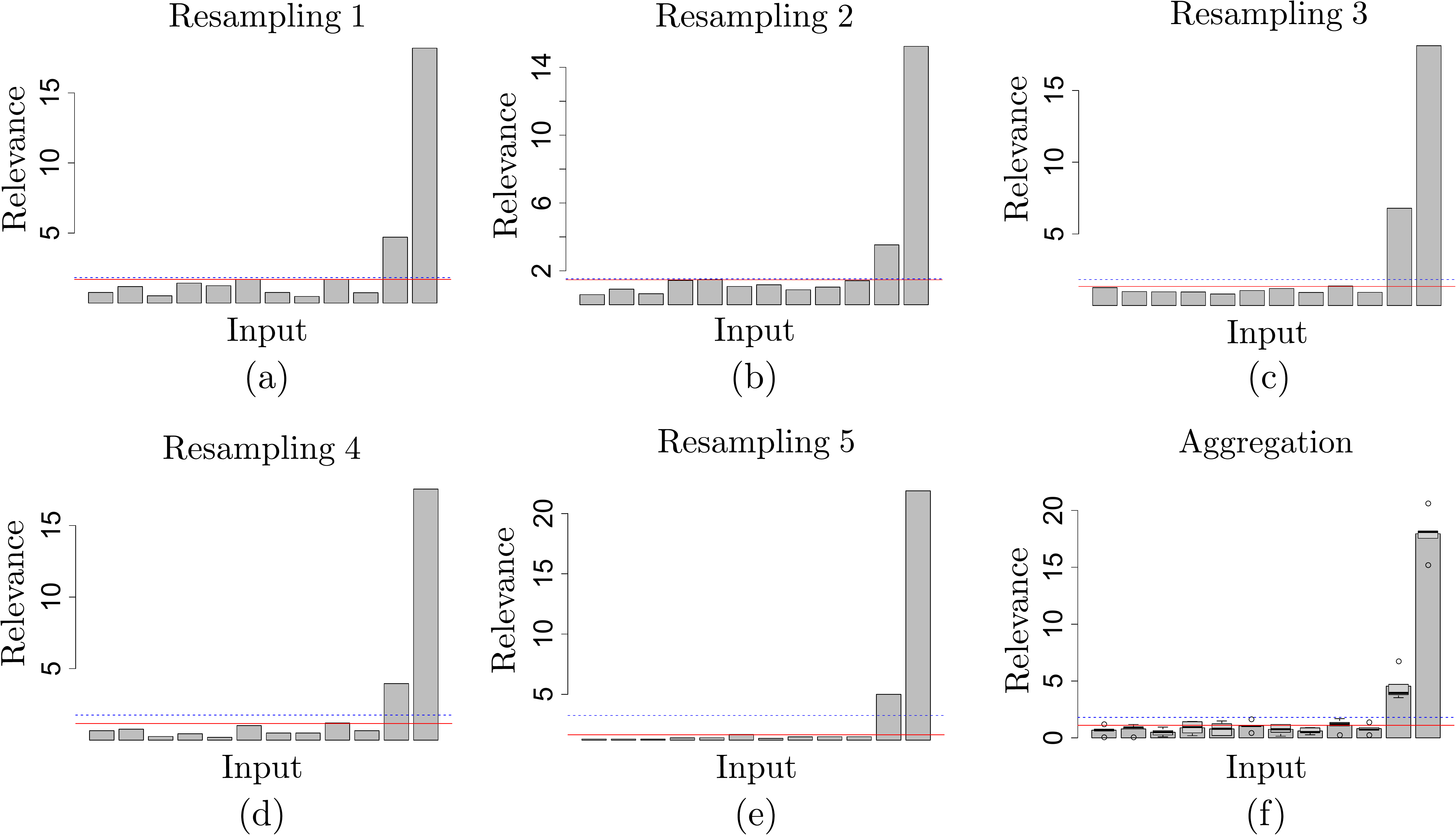}
\caption{\textbf{(a--e)} Results of five resamplings for the Logistic map indicate the embedding parameter (2, 1). \textbf{(f)} Aggregation of the five resamplings. The maximum embedding dimensions, or MEB, were set as ${(m_{\max} = 5, \tau_{\max} = 3)}$.}
\label{fig:logistic}
\end{figure}

We have tested that our approach yields consistent results for different datasets and different initializations, reinforcing that a stable pattern is being learned. \figref{logistic}(a-e) show the relevances while running the network for five resamplings on the Logistic map. In this circumstance, we considered the search space provided by (${m_{\max}=5,\tau_{\max}=3}$).

\figref{logistic}(f) shows the average of the five resamplings. Here and next, box plots~\citep{mcgill:78} are drawn on each bar to indicate the variance of relevances along resamplings. Very similar results were obtained for other datasets (not included in this manuscript to avoid redundancy), and the aggregated plot is used to estimate the embedding parameters. By analyzing \figref{logistic} while using the threshold procedure outlined in \secref{visual-inspection-of-ps-parameters}, we observe that all resamplings suggest, with high confidence as made evident by narrow box plots, an embedding dimension $m=2$ and time delay $\tau = 1$, matching the ground truth as desired (\tabref{comparison-results}). 

\begin{figure}[ht]
\centering
\includegraphics[width=.90\linewidth]{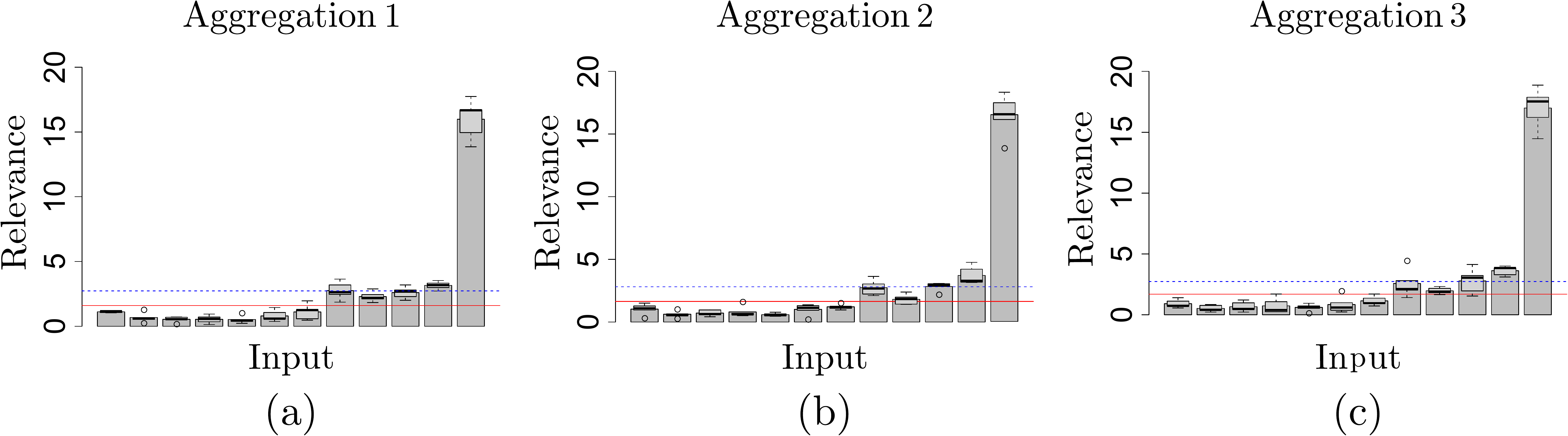}
\caption{Results for the H\'enon map under three different initializations. Respective estimatives are \textbf{(a--c)}: $(2, 4)$, $(3, 4)$, $(3, 4)$.}
\label{fig:henon}
\end{figure}

In order to reinforce the robustness of our method with respect to network initialization, we performed three experiments with the H\'enon map, using three different random strategies, as outlined in \secref{datasets}. In all situations, we defined the search space using $(m_{\max}=4,\tau_{\max}=4)$.
\figref{henon} shows the plots of aggregated importance for these experiments. One may notice very similar relevances and almost the same embedding parameters, regardless the initialization.

\subsection{Lorenz: Consistency along the search space}
\label{sec:lorenz}

Lorenz is produced from a nonlinear system which is more complex than the Logistic and the H{\'e}non maps. This dataset is used to study the robustness of our method with respect to variations in the search space. In this sense, as the search space in our case is represented by the number of input neurons, we ran our network under different MEB parametrizations $(m_{\max},\tau_{\max})$ and analyzed how predictions ($m,\tau$) varied under such conditions. All other network settings remained the same as discussed in \tabref{network-settings}.

\begin{figure}[h!]
\centering
\includegraphics[width=.85\linewidth]{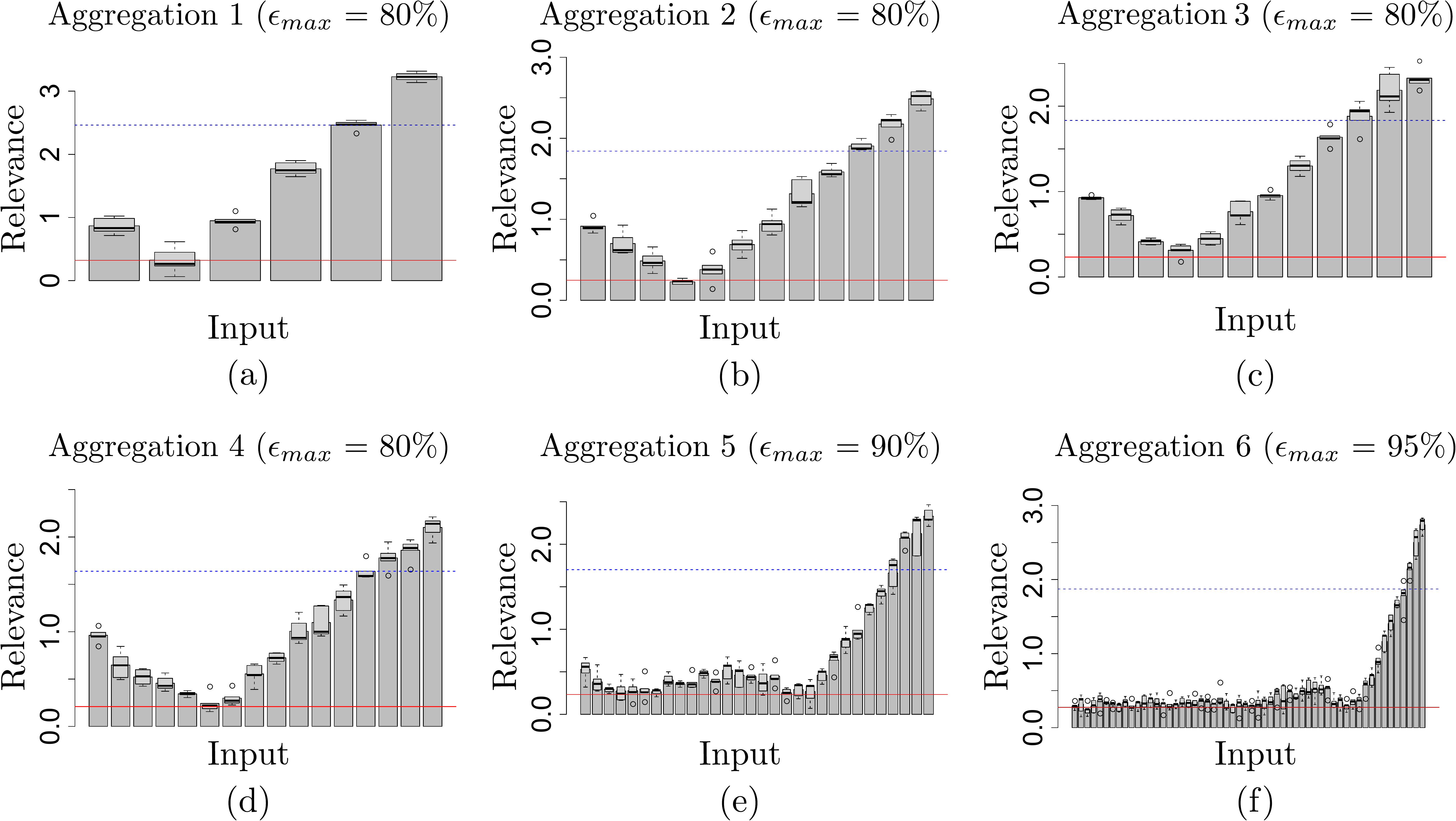}
\caption{Robustness of the estimation of embedding parameters as function of the initial search spaces $(m_{\max},\tau_{\max})$ for the Lorenz system. From \textbf{(a--f)}, MEB are: $(3,3)$, $(5,3)$, $(7,2)$, $(3,8)$, $(6,6)$, $(8,8)$. Respective estimated parameters: $(2,5)$, $(3,8)$, $(3,9)$, $(3,11)$, $(3, 11)$, $(3,13)$.}
\label{fig:lorenz}
\end{figure}

The experiment results, shown in \figref{lorenz}, reinforce that excessively small MEB values may create a network whose architecture is not big enough to capture the system dynamics (\figref{lorenz}(a)). Conversely, similar values of embedding parameters can be estimated when smaller-to-medium values of MEB are used (\figref{lorenz}(b-d)). On the other hand, by excessively increasing the search space, it is more difficult to find a clear set of parameters $(m, \tau)$ as the model captures more disturbances especially in nonlinear systems such as Lorenz. In such cases, in attempt to obtain a highly confident estimation for $(m,\tau)$, one needs to increase the threshold $\epsilon_{\max}$ from our model, as illustrated in \figref{lorenz}(e,f), where we have increased the upper threshold $\epsilon_{\max}$ to $90\%$ and $95\%$, respectively.

The experiment also suggests that the range of MEB is an important parameter, but not crucial (in the sense that we can deal with ``acceptable'' variations of it) for the estimation of embedding parameters. After applying our method for smaller-to-greater values of MEB, we can see that the network architecture led to similar patterns specially in middle-range values (\figref{lorenz}(b-d)). This goes in accordance to the Bias-Variance Dilemma, which states that one should choose an algorithm bias that is not too restricted (prone to underfitting) nor too relaxed (where complex functions will tend to overfit/memorize the data). For general systems, we suggest at first to use typical (according to the related work) values of MEB that lead to ${|\text{PEV}| - 1 = [12, 30]}$ input neurons.

In addition, as the network was trained using a different number of inputs (maximum embedding bounds) and its architecture still led to similar outputs of $m$ and $\tau$, this experiment suggests that even using different embeddings, the neural network is robust enough to converge to the Lorenz dynamics (\figref{embeddings-example}(b)). This goes in accordance the claiming that $m$ and $\tau$ are bounded by the time delay window $t_w$, and that several tuples ($m,\tau)$ can be used to unfold the attractor.

\subsection{R\"ossler: Forecasting accuracy}
\label{sec:rossler}

\begin{figure}[ht]
\centering
\includegraphics[width=.65\linewidth]{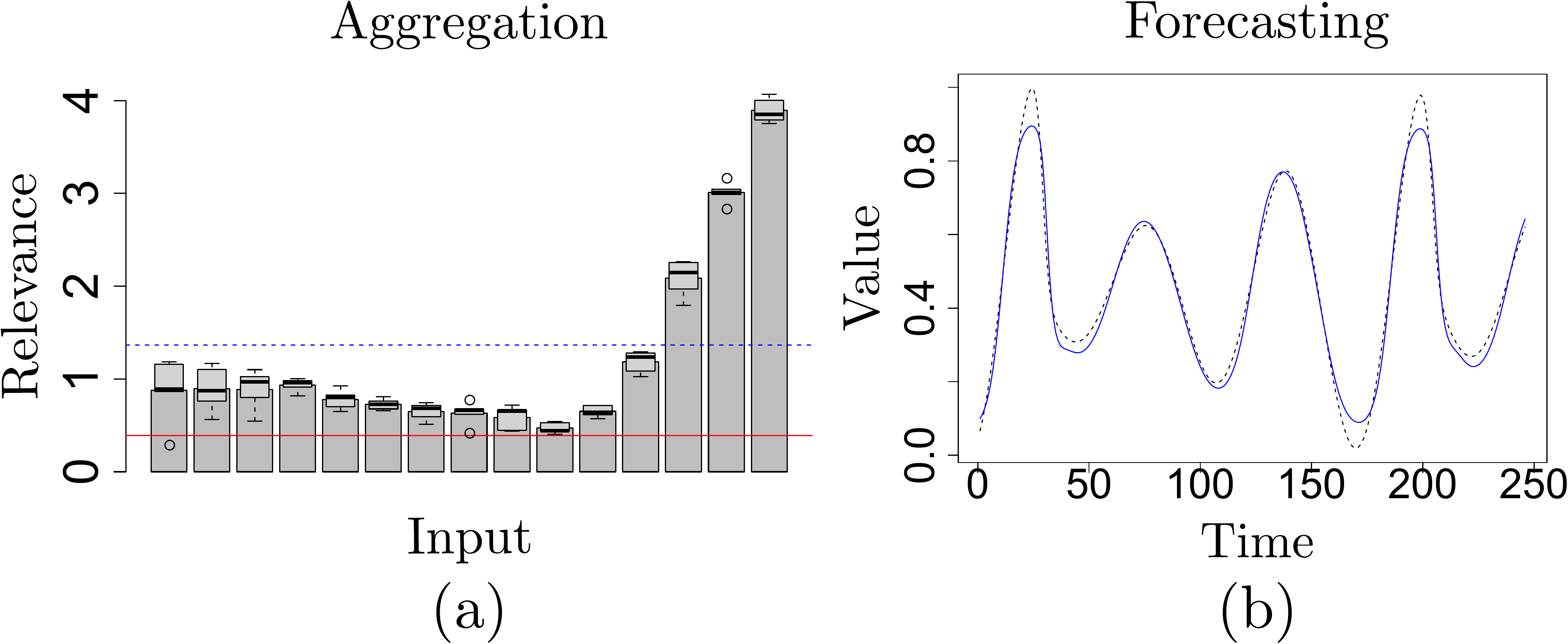}
\caption{Results for the R\"ossler system. \textbf{(a)} Relevance of input neurons. \textbf{(b)} Comparison of the $250$ forecasted (solid blue) and expected (dashed red) observations.}
\label{fig:rossler}
\end{figure}

Besides comparing the estimated embedding parameters with known ground truth, a different way of assessing the performance of the proposed neural network is by \emph{predicting} data. We conducted such strategy using the R\"ossler dataset, another well-known benchmark in the context of Dynamical Systems~\citep{rossler:76:physics}. Starting with an initial search space set in form ${(m_{\max}=4,\tau_{\max}=5)}$, we obtained the embedding parameters $(m=3,\tau=6)$ as shown in \figref{rossler}(a). We refer to \secref{visual-inspection-of-ps-parameters} for details about the blue and the red lines defining upper and lower bounds to support the selection of embedding parameters.

Complementary, \figref{rossler}(b) shows the predicted (solid blue) \emph{vs} the expected (dashed black) series for a \emph{single} observation forecasting under $250$ time steps. The image shows the forecasting using the last $k$-folded network. As it can be seen, the experiment suggests the network was capable to reveal the dynamics of the dataset.


\subsection{Sunspot and Gaussian Distribution: Analyzing real-world and noise data}
\label{sec:sunspot}

\begin{figure}[htb]
\centering
\includegraphics[width=.95\linewidth]{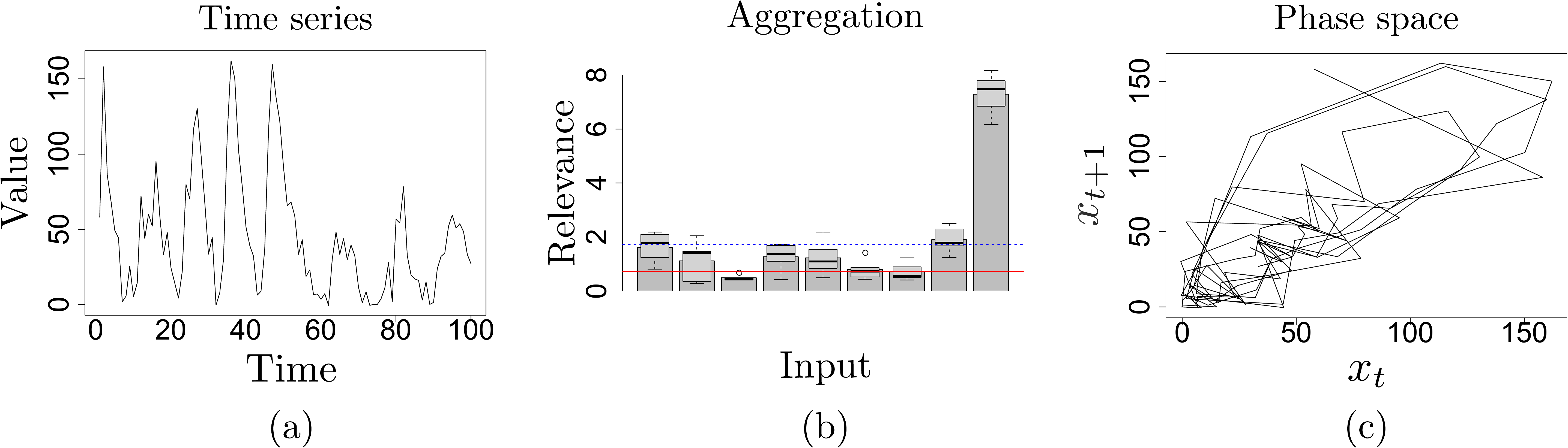}
\caption{\textbf{(a)} Sunspot time series. The series has a lot of fluctuations that jeopardizes the visualization of the attractor (we see a lot of concentric ellipses overlapping each other as approaching the origin, make a blur of points difficult to see its structure). Therefore, we have proceeded with a smoother version, a spline curve made of $100$ equally-spaced observations ($10\%$) from the original dataset. \textbf{(b)} Relevance of dimensions. \textbf{(c)} Phase space plot with $(m=2,\tau=1)$.}
\label{fig:sunspot}
\end{figure}

In our last experiment, we evaluated the effectiveness of our method on the Sunspot series~\citep{andrews:85:book}, a dataset formed with real-world observations, having a fragment of it illustrated in \figref{sunspot}(a). In this situation, nothing is known about the series generating rule $R(\cdot)$ and no ground truth is available for assessing the quality of the estimated embedding parameters. In those scenarios, one can rely on the visual analysis and properties of both time series and  embeddings (only seeing the first two or three dimensions of it) in attempt to validate the parameter estimation by their similarities to other well-known datasets.

Using our network approach on Sunspots with an initial search space $(m=4,\tau=3)$, we found the embedding parameters $(m=2,\tau=1)$ (\figref{sunspot}(b)). This estimation is also reinforced by the fact that the Sunspot dataset (\figref{sunspot}(a)) resembles sinusoidal characteristics, such as in \figref{sine-example}. Finally, \figref{sunspot}(c) shows the attractor for the phase space estimated using our approach. Indeed, this figure shows several concentric sharped-elliptic-like trajectories, as expected from some series composed of a sum of functions with sinusoidal trends.

\begin{figure}[ht]
\centering
\includegraphics[width=.45\linewidth]{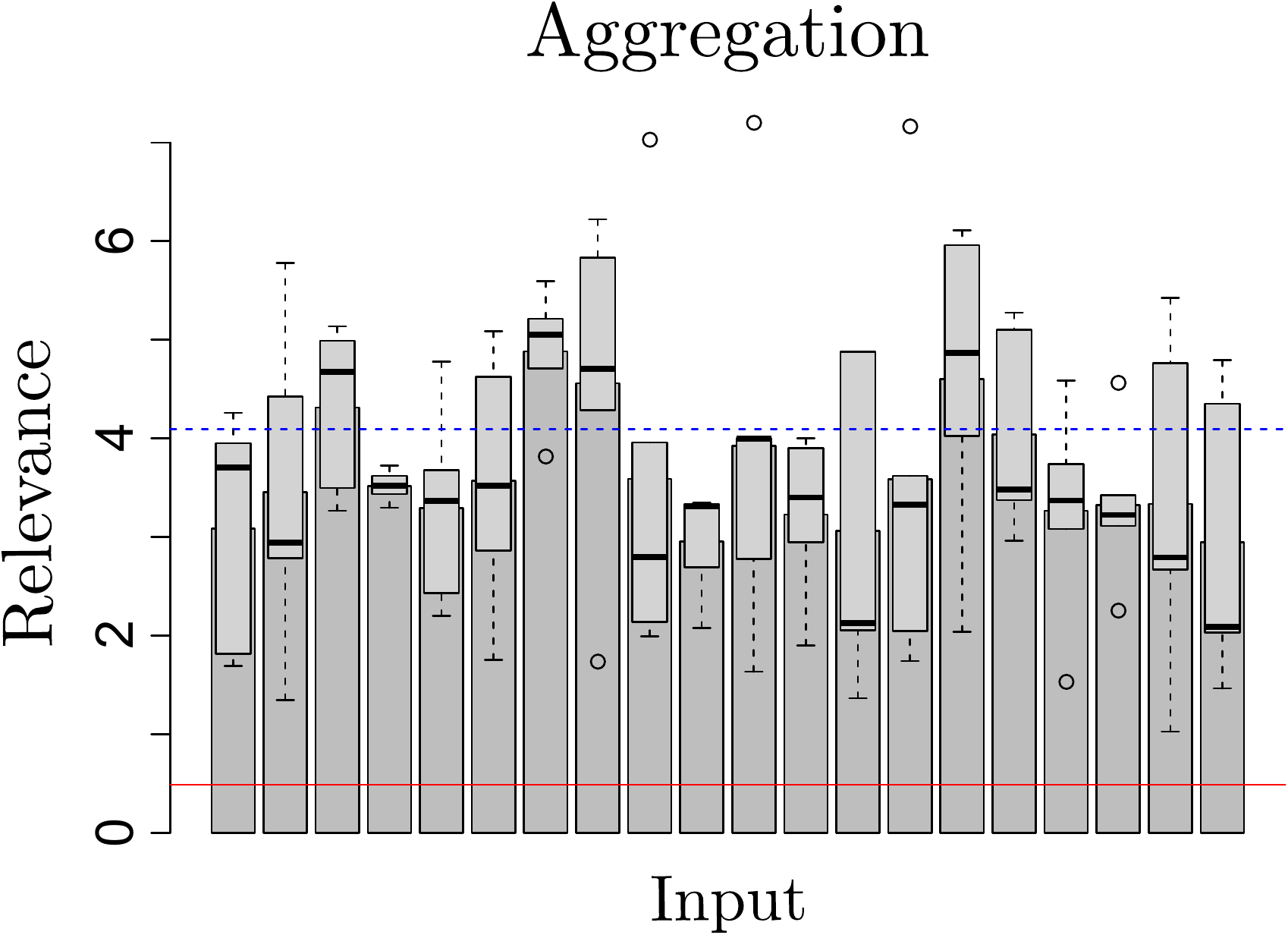}
\caption{Relevance of dimensions for data produced using the Gaussian distribution $\fold{G}(\mu = 0, \sigma = 1)$. There is no evident manner to select the embedding parameters $(m,\tau)$ in this specific scenario.}
\label{fig:gaussian}
\end{figure}

As consequence of analyzing real-world datasets, we also consider a pure-randomly generated time series following a Gaussian distribution ${\fold{G}(\mu = 0, \sigma^2 = 1)}$, where $\mu$ and $\sigma$ correspond to the mean and the standard deviation, respectively. Here, we estimated embedding parameters using an initial search space defined as ${(m_{\max}=5,\tau_{\max}=5)}$. \figref{gaussian} shows the results. As one may notice, there is no trivial way to select a subset with the most relevant dimensions (which would provide $m$), nor a manner to point out a minimum below $\epsilon_{min}$ (which would give us $\tau$). Moreover, the variance of relevances is very large for most dimensions, what goes in accordance with Chaos Theory~\citep{alligood:1996}. In those circumstances, it is expected that the phase states of stochastic series do not form any structure, but rather fully spread all over the embedding space in some hyperspherical organization~\citep{mello:book:18}, such that $m$ is always equal to the maximum embedded dimension.

\begin{figure}[ht]
\centering
\includegraphics[width=.95\linewidth]{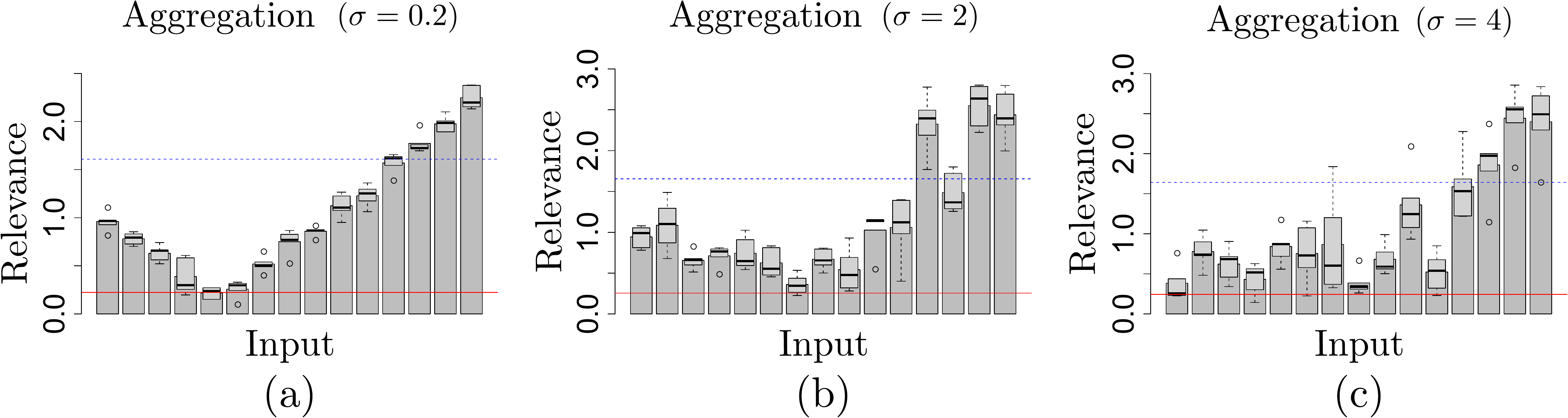}
\caption{From \textbf{(a)} to \textbf{(c)}, our model estimated $(m = 3, \tau = 10)$, $(m = 2, \tau = 3)$ and, $(m = 3, \tau = 5)$ after adding $\fold{G}(0, \{0.2,2,4\})$ to the Lorenz system.}
\label{fig:lorenz-noise}
\end{figure}

As a last experiment, we test the robustness of our method after adding Gaussian noises with $\mu = 0$ and $\sigma =\{0.2,2,4\}$ to the Lorenz system (similar results were obtained for other datasets), using a network with MEB ${(m_{\max}=4,\tau_{\max}=5)}$. \figref{lorenz-noise} illustrates the results. For relatively low amount of noise ($\sigma = 0.2$), our method is still capable of recovering the phase-space dynamics, finding ($m =3, \tau = 10$) as embedding parameters, as shown in \figref{lorenz-noise}(a). As the signal to noise ratio decreases, \emph{i.e.}, as the amount of noise increases, the estimation got twisted (as expected), leading to a estimation of (${m = 2, \tau = 3}$) and (${m = 2, \tau = 5}$) for $\sigma = 2$ and $\sigma = 4$, respectively. Additionally, it is also worth to mention that this problem leads to different inconsistencies when compared to variations on the search space (\secref{lorenz}). There, even when too much dimensions were involved in the training, the variance on the box plots remained low for most of the dimensions. Here, the opposite scenario is observed: box plots show great variations in their quantiles even for few dimensions. Therefore, this experiment also shows that box plots are not just useful to show if the network has converged to a solution, but also to qualitatively measure the amount of randomness in the time series. In such context, estimations from \figref{lorenz-noise}(b) and \figref{lorenz-noise}(c) are not trustworthy due to high variances over dimension relevances. Moreover, in those cases, its better to first filter the dataset to later proceed to further analyses.

\section{Conclusions and Future Work}
\label{sec:conclusion}

Several statistical approaches from the literature support time-series analyses, specially in terms of forecasting~\citep{box:15:book}; however they are insufficient to deal with complex and chaotic data. Dynamical Systems tackle such a problem by reconstructing time series into phase spaces, unveiling the relationships among observations, consequently leading to more consistent models. In this context, methods have been proposed to support the reconstruction of phase spaces by estimating the optimal set of embedding parameters $m$ and $\tau$, both necessary to unfold time dependencies as provided by Takens' embedding theorem~\citep{takens:1981}. As main drawback, those methods rely on predefined measurements to compare different phase spaces and estimate the most adequate after analyzing a set of possibilities.

As an alternative, we proposed the usage of an artificial neural network with a forgetting mechanism to implicitly learn the embedding parameters while mapping input examples to their expected outputs. Despite small similarities with MC~\citep{manabe:neuro:07}, tour approach is much more simpler as it does not require hidden unit clarification, selective learning nor pruning heuristics during training. The single parameter our approach requires is the maximum embedding bound (MEB), which is used to define the length of the input layer, thus satisfying R1 (see \secref{introduction}). Moreover, we rely on a different normalization of initial weights, as well as another criterion to define relevant dimensions, thus positively impacting the estimations of $m$ and $\tau$.

It is worth to reinforce that we relied on the PEV format (\eqnref{pev}) to built our neural network not because we were \emph{just} worried about the final learned model, as is the case of several neural networks proposals, but also to derive embedding the parameters from the final skeletal architecture of the network. For that, we based on the distribution of relevant neurons to estimate $m$ and $\tau$. Therefore, if we had used other feature vector as input for our neural network (\emph{e.g.} PCA~\citep{jolliffe:1986} over an arbitrary overestimated embedding), extracting back $m$ and $\tau$ from the final architecture would not be trivial, perhaps not possible.

We have also performed experiments to assess the sensitivity of our approach to different random initializations and search space settings. As made evident throughout the experiments, our method achieved robust and consistent results for several datasets and MEB values (also reinforced by small variances of relevances of the dimensions along resamplings), satisfying R3 and R2, respectively. As future work, we intend to tackle the butterfly effect by proposing a network to output recursive forecasting, perhaps by means of Recurrent Neural Networks or more sophisticated methods.

\section*{Acknowledgements}
We acknowledge sponsorships of FAPESP (S\~{a}o Paulo Research Foundation) and CNPq (National Counsel of Technological and Scientific Development).  Any opinions, findings, and conclusions or recommendations expressed in this material are those of the authors and do not necessarily reflect the views of FAPESP nor CNPq.

\section*{Disclosure statement}
The authors declare that they have no conflict of interest.

\section*{Funding}
This research was supported by FAPESP and CNPq, grants 2018/10652-9, 2017/16548-6 and 302077/2017-0.

\bibliographystyle{abbrvnat}
\setcitestyle{authoryear}
\bibliography{bibliography}

\end{document}